\documentclass[letterpaper, 10 pt, conference]{ieeeconf}  % Comment this line out if you need a4paper

\usepackage{amssymb,times}
\usepackage{bbm}
\usepackage{cite}
\usepackage{verbatim, amsfonts,enumerate}     
\usepackage{bm,array}
\usepackage{graphicx}
\usepackage{subfigure}
\usepackage[cmex10]{amsmath}
\usepackage[font=normalsize]{caption}
\usepackage{algorithm,algorithmic}
\usepackage{color,url}
\usepackage[table]{xcolor}
\usepackage[normalem]{ulem}

\input{mysymbol.sty}

\newtheorem{example}{\bf Example}
\newcommand{\INDSTATE}[1][1]{\STATE\hspace{3mm}}
\newcommand{\INDSTATED}[1][1]{\STATE\hspace{6mm}}

\usepackage{amsmath}
\usepackage{amsfonts}
\usepackage{threeparttable}

\newcommand{\inclu}[0] {\ar@{^{(}->}}

\newcommand{\EE}{{\mathbb E}}

\newcommand\blfootnote[1]{%
	\begingroup
	\renewcommand\thefootnote{}\footnote{#1}%
	\addtocounter{footnote}{-1}%
	\endgroup
}

\title{Dealing with Sparse Rewards in Continuous Control Robotics via Heavy-Tailed Policies}

\author{Souradip Chakraborty$^{1}$, Amrit Singh Bedi$^{1}$, Alec Koppel$^{2}$, Pratap Tokekar$^{1}$, Dinesh Manocha$^{1}$% <-this % stops a space
}

\begin{document}

\maketitle
%\thispagestyle{empty}
%\pagestyle{empty}

%%%%%%%%%%%%%%%%%%%%%%%%%%%%%%%%%%%%%%%%%%%%%%%%%%%%%%%%%%%%%%%%%%%%%%%%%%%%%%%%
\begin{abstract}
In this paper, we present a novel Heavy-Tailed Stochastic Policy Gradient (HT-PSG) algorithm to deal with the challenges of sparse rewards in continuous control problems.  Sparse reward is common in continuous control robotics tasks such as manipulation and navigation, and makes the learning problem hard due to non-trivial estimation of value functions over the state space. This demands either reward shaping or expert demonstrations for the sparse reward environment. However, obtaining high-quality demonstrations is quite expensive and sometimes even impossible. We propose a heavy-tailed policy parametrization along with a modified momentum-based policy gradient tracking scheme (HT-SPG) to induce a stable exploratory behavior to the algorithm. The proposed algorithm does not require access to expert demonstrations. We test the performance of HT-SPG on various benchmark tasks of continuous control with sparse rewards such as 1D Mario, Pathological Mountain Car,  Sparse  Pendulum in OpenAI Gym, and Sparse MuJoCo environments (Hopper-v2). We show consistent performance improvement across all tasks in terms of high average cumulative reward. HT-SPG also demonstrates improved convergence speed with minimum samples, thereby emphasizing the sample efficiency of our proposed algorithm.
\end{abstract}

%%%%%%%%%%%%%%%%%%%%%%%%%%%%%%%%%%%%%%%%%%%%%%%%%%%%%%%%%%%%%%%%%%%%%%%%%%%%%%%%%%%%%%%%%% S  E  C  T  I  O  N %%%%%%%%%%%%%%%%%%%%%%%%%%%%%%%%%%%%%%%%%%%%%%%%%%%%%%%%%%%%%%%%%%%%%%%%%%%%%%%%%%%%%%%%%%%%%%%%%%%%%%%%%%%%%%%%%%%%%%%%%%%%%%%%%%%%%%%%%%%%%%%%%%%%%%%%%%%%%%%%%%%%%%%%%%%%%%%%%%%%%%%%%%
\section{Introduction}
Reinforcement learning (RL)\blfootnote{This research was supported by Army Cooperative Agreement W911NF2120076 and ARO grant W911NF2110026.
	\\ S. Chakraborty, A. S. Bedi, P. Tokekar, and D. Manocha are with the University of Maryland, College Park, MD, USA. Email: \{schkra,amritbd,tokekar,dmanocha\}@umd.edu. A. Koppel is with JP Morgan Chase AI Research, NY, USA. Email: \{{aekoppel314}@gmail.com\}.}
%is a paradigm in which an autonomous agent or robot interacts with the environment, receives corresponding rewards in terms of feedback, and then learns to take optimal actions by maximizing the cumulative rewards \cite{sutton1998reinforcement}. RL has enjoyed huge recent successes in different applications such as Atari games \cite{mnih2015human,van2016deep}, AlphaGo in Go \cite{silver2016mastering}, finance \cite{charpentier2021reinforcement}, etc. 
%RL
has been employed with great success in several  continuous control robotic tasks such as grasping \cite{kilinc2021reinforcement}, motion planning \cite{everett2018motion}, and navigation \cite{liu2020robot}. The key underlying idea in RL is to explore in an unknown environment, collect rewards, and then move to maximize the reward collection. In the real world, designing dense rewards is challenging for robotic tasks such as manipulation and navigation \cite{8957584}. Reward engineering for robotic tasks is difficult due to complex state space representations and usually requires manually-designed perception systems of
the environment \cite{singh2019end}. Hence, it makes sense to work directly with naturally specified sparse rewards \cite{9341714,kang2018policy,rengarajan2022reinforcement}. For example, it is much easier to specify a binary reward ($1$ for successful completion of a task and $0$ otherwise) than to come up with a dense reward structure. However, learning with sparse rewards is much more challenging because it results in the Hessian of the value function with respect to policy parameters being ill-conditioned. It also imposes the need to sample multiple trajectories in order to have a nontrivial estimate of the value function, which is sample inefficient \cite{rauber2021reinforcement}.

\begin{figure}[t]
	\centering
\subfigure[Sparse Inverted Pendulum.]{		\includegraphics[width=0.4\columnwidth]{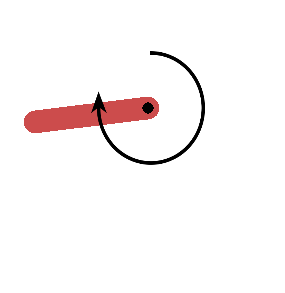}}
\subfigure[Sparse Hopper-v2.]{		\includegraphics[width=0.4\columnwidth]{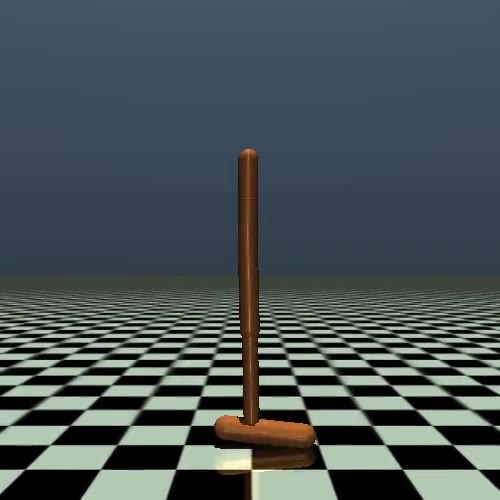}}
	\caption{\normalsize Sparse reward continuous control robotic environments. (a) Sparse Inverted Pendulum task of OpenAI Gym \cite{brockman2016openai}. The state-space includes position of the free-end of the Pendulum in Cartesian coordinates $(x,y)$ and velocity. There is only one continuous action that represents the angular torque $\in[-2,2]$. A non-zero reward is given only when the agent reaches a specific angle (from $-2$ to $2$ degrees), which is an instance of sparse reward. (b) One-legged hopper from Hopper-v2 environment in MuJoCo. This is a continuous control robotic task with $12$-dimensional state space and $3$-dimensional action space. The goal is to stand for as long as possible and the episodes end when the hopper fell over, which is defined by thresholds on the torso height and angle. A reward of $+1$ is provided only after the agent moves forward over $2$ units from its initial position. The reward here is also sparse in nature.}
	\label{Hopper}
	\end{figure}

Furthermore, learning from sparse rewards in continuous control robotic tasks becomes even more challenging (as mentioned in Fig. \ref{Hopper}) because they exhibit continuous state and action spaces. For instance, in manipulation tasks, joint angles of robots are continuous, and in navigation tasks, the pose of robots and control inputs are continuous.  RL in continuous control problems is hard because it's hard to compute expectations with respect to continuous state distributions and continuous actions to evaluate value functions \cite{van2007reinforcement}. 

The issue of sparse rewards is usually dealt with in literature either through either reward shaping \cite{botteghi2020reward,mataric1994reward,reward_1} or utilizing expert demonstrations \cite{vecerik2017leveraging,ng2000algorithms,ziebart2008maximum,kang2018policy,libardi2021guided}. Intuitively, both of these approaches try to induce effective exploration into the sparse reward environment by providing surrogate rewards. Reward shaping approaches modify the reward feedback to motivate the agent to visit unexplored states in the environment. For instance, authors in \cite{reward_4}  induce such behaviors via intrinsic curiosity,  and \cite{reward_2} utilizes information to motivate the exploration.  Another line of work utilizes expert's demonstrations to learn effectively in sparse reward environments \cite{reward_1,hester2018deep,kang2018policy,rengarajan2022reinforcement}. The main idea here is to either use available demonstration to clone an expert's behavior (imitation learning) or just utilize demonstrations to provide additional rewards to guide the exploration \cite{kang2018policy,rengarajan2022reinforcement}. But the major limitation of these approaches depends on the quality of the expert demonstrations. If the demonstrations are not sub-optimal or not good, these approaches fail badly. Apart from that, obtaining a high-quality demonstration is quite expensive, especially in robotic environments \cite{kilinc2021reinforcement}.

In contrast to existing approaches to deal with sparse reward settings, in this work, we follow a different route and take motivation from the global convergence results in tabular MDP settings \cite{mei2020global}. A crucial enabler for learning global optimal policies in \cite{mei2020global} is the idea of \emph{persistent exploration}, which helps to implicitly induce sufficient exploration in the state space. This ensures that the probability of taking any action in a given state is always non-zero, which would help to visit the complete state space and look for rewards. 
%But in this work, we are interested in robotic environments which are mostly continuous in state and action spaces, and hence similar ideas would create issues because many common distributions in continuous space may fail to be integrable if their likelihood is lower bounded away from null over the entire (not necessarily compact) state space.
Recently, authors in \cite{bedi2021sample} have extended the idea of \emph{persistent exploration} to continuous spaces and have proposed to utilize heavy-tailed policies to avoid convergence of policy gradient methods to spurious local maximas. Taking motivation from \cite{bedi2021sample}, we ask the following question
%behavior and remains stuck in the state space. 
%This is also important because otherwise, with the existing policy distributions (in the literature)  such as Gaussian policies exhibit a poor exploration across the state action space, and often settles to spurious behavior and remains stuck in the state space.  Hence,  we ask the following question:

``\emph{Can heavy-tailed policies make model-free RL sample efficient for practical robotics tasks that involve sparse reward structure without any expert demonstrations?}"

We answer this question in affirmative in this paper and propose to utilize heavy-tailed policies (such as Cauchy) for policy parametrization along with a modified momentum-based policy gradient tracking to deal with the sparsity in reward. These heavy-tailed distributions appear heavily in fractal geometry\cite{hutchinson1981fractals,mandelbrot1982fractal},  finance\cite{taleb2007black,taylor2009black}, pattern formation in nature \cite{avnir1998geometry}, and networked systems \cite{clauset2009power}, but has not been well investigated in RL framework. Intuitively, heavy-tailed policies induce an implicit exploration behavior into the trained policies (because of the high probability of taking tail actions), and help to learn effectively in sparse environments even without any expert demonstrations. We summarize the main contributions of this paper as follows.

\begin{itemize}
    \item We propose a novel way to deal with sparse reward environments to train a policy in continuous state-action space environments. Our approach is fundamentally different because we introduce heavy-tailed policy parametrization and avoid using expert demonstrations, which is a common practice in the existing literature. This provides a way to work with sparse reward environments without any reward shaping or demonstrations, which is very difficult otherwise. Additionally, our formulation is flexibly designed to efficiently incorporate prior demonstrations as well, if available.
    
    \item We observed that just replacing Gaussian policy parametrization with heavy-tailed (Cauchy) parametrization results in unstable behavior during the training. Hence, we propose a modified version of the momentum-based tracking method proposed in \cite{cutkosky2019momentum} to control the variance of the stochastic gradient estimates. 
    
    \item Finally, we show the efficacy of the proposed algorithm on various continuous control task problems. The proposed algorithm shows consistent performance improvement over a variety of benchmark problems (cf. Sec.~\ref{sec:experiments}). 
\end{itemize}

% \textbf{Motivating Example:} To emphasize the importance of learning in a sparse reward continuous control robotic environment, consider the problem of one-legged Hopper in MuJoCo environments shown in Fig.~\ref{Hopper}. It's easier to define sparse reward for these continuous tasks, rather than providing reward feedback after some time for each particular action. 
% \subsection{Related Work}

\noindent \textbf{Reward Shaping:} Reward shaping is the most intuitive way to deal with sparse rewards. The idea was first appeared in  \cite{mataric1994reward} and further developed in recent works \cite{reward_1,reward_2,reward_3,reward_4}. The main idea revolves around intrinsic curiosity \cite{reward_4} and information gain based shaping \cite{reward_2}. Besides being simple, these methods come with the challenge of designing the additional reward functions which require expert supervision and demonstrations which are expensive. Additionally, it also induces expert-specific bias to the learning systems which ultimately leads the agent to explore only certain parts of the environment hindering the overall improvement.

\vspace{2mm}
\noindent\textbf{Imitation Learning:} Another line of work focuses on cloning an expert behavior called imitation learning (IL) \cite{hussein2017imitation}. Inverse reinforcement learning (IRL) is one way to do IL by extracting rewards from the given set of expert's trajectories for a given task \cite{ng2000algorithms,ziebart2008maximum}. This issue of reward estimation was resolved by generative adversarial imitation learning (GAIL) algorithm by utilizing a discriminator to provide reward functions \cite{ho2016generative}. But the main drawback of IL-based approaches is that they do not utilize the feedback from the environment and behave according to the policies learned from demonstrations. Our approach in this work is fundamentally different, and we propose a method that works without demonstrations and can also incorporate prior demonstrations efficiently in the methodology, is available.

\vspace{2mm}
\noindent\textbf{Learning from Demonstration:} The idea here is to utilize expert's demonstrations to guide the standard learning procedure in RL algorithms
\cite{schaal1996learning,kang2018policy,libardi2021guided,rengarajan2022reinforcement,chen2020policy}. Authors in \cite{hester2018deep,rajeswaran2017learning} proposed to include expert demonstrations to replay buffers and utilize them to accelerate the learning. The authors in \cite{kang2018policy} proposed an effective way to combine information from expert's policy to guide the exploration in the policy gradient algorithms. Mainly, the original reward function is modified to also include a term that accounts for the distance of current policy to the expert's policy. But as mentioned previously, the major drawback here is also their dependence upon the availability of demonstrations, which are hard to get in practice for continuous control problems. For instance, expert's demonstrations in \cite{rengarajan2022reinforcement} are obtained by running TRPO with dense rewards and then later used to train a policy with sparse rewards in the same environment. This could be difficult to achieve in practice. Therefore, we propose to modify the policy parametrization in continuous control environments to induce the required exploration in the learning procedure. 

\textbf{Heavy-Tailed Policy Parametrization:} The idea of parametrizing policies via heavy-tailed distribution has appeared in the reinforcement learning literature \cite{chou2017beta,bedi2021sample}. The authors in \cite{chou2017beta} proposed to utilize beta distribution for policy parametrization but are restricted to dense reward structure environments. Authors in \cite{bedi2021sample} have focused on the development of heavy-tailed policy gradient to avoid convergence to local maxima and do not explicitly deal with sparse rewards. This work focus on sparse reward continuous control environments and extensive experimental evaluations to support the importance of heavy-tailed policy parametrization.

The paper is organized as follows. We start with  the problem formulation in Sec.~\ref{sec:prob}, followed by proposed algorithm in Sec.~\ref{sec:proposed}. We present experimental results in Sec.~\ref{sec:experiments}, and then conclude the paper in Sec.~\ref{sec:conclusion}.

\section{Markov Decision Problems with Sparse Rewards}\label{sec:prob}
\begin{figure}[t]
	\centering
	\subfigure[1D Mario environment \cite{matheron2019problem}.]{\label{fig:b}		\includegraphics[width=0.5\columnwidth]{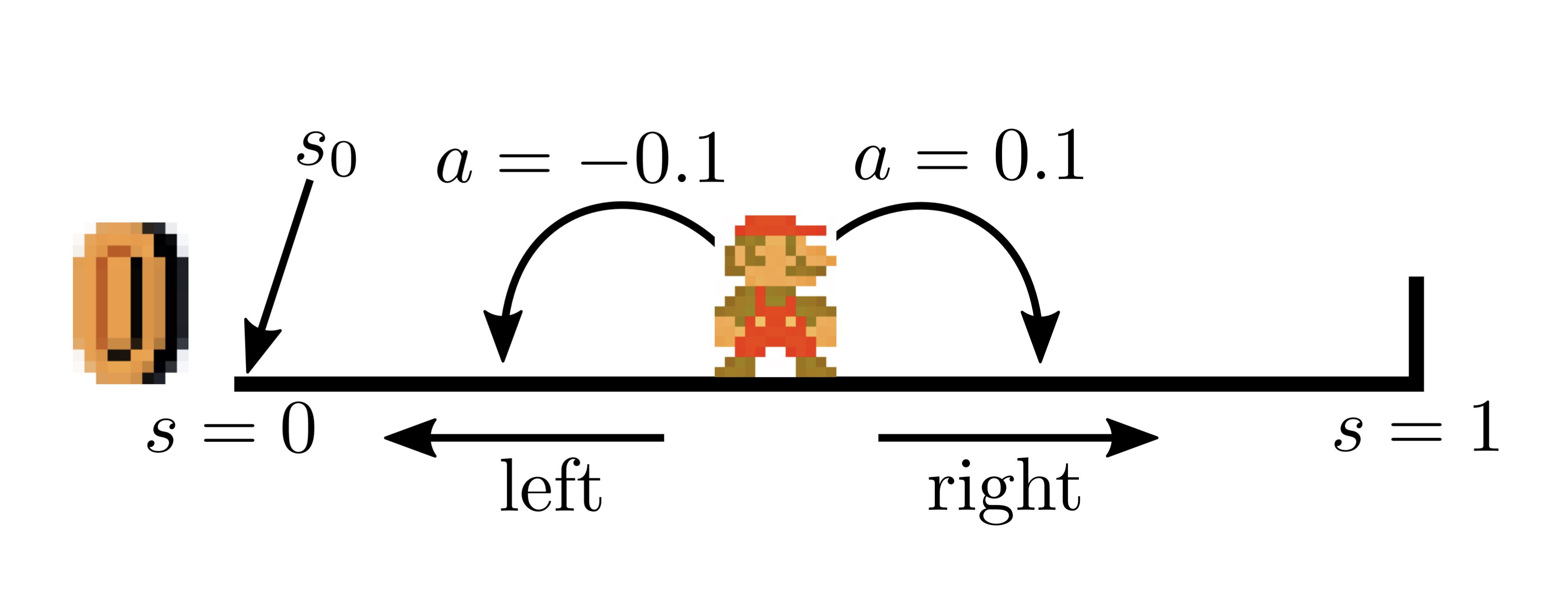}}
	\subfigure[PMC.]{\label{fig:a}		\includegraphics[width=0.4\columnwidth]{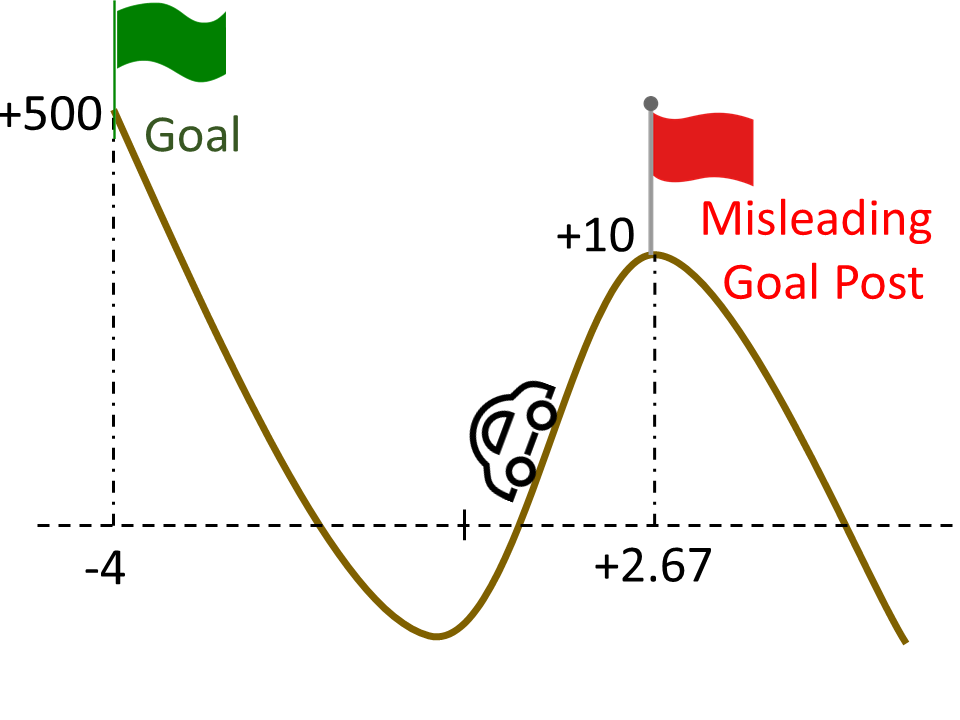}}
	\caption{\normalsize Sparse reward continuous control environments. (a) 1D Mario environment where the goal is to collect coin placed at $s=0$. A reward of $1$ is provided when Mario reaches $s=0$, otherwise no reward for taking any action in the environment. (b) Pathological Mountain Car where the goal is to reach top of the hill. This is an instance where long and short-term incentives are misaligned in continuous space. There is a low reward state (red) and another high reward (red) state atop a higher hill. Policies that do not incentivize exploration get stuck at the spurious goal.}
		\label{fig2}
	\end{figure}

When we formulate the continuous control robotics problems via reinforcement learning (RL), an autonomous robot interacts with the underlying environment by visiting different states in the state space $\mathcal{S}$. It starts from a particular state $s\in\mathcal{S}$,  selects an action $a\in\mathcal{A}$ from the action space, and then transitions to another state $s'\in\mathcal{S}$ in the state space. The next state is assumed to follows an unknown Markov transition density $\mathbb{P}(s'|s,a)$. Then after reaching state $s'$, agent received an instantaneous reward of $r(s,a)$ which quantifies the merit of decision $a$ at state $s$. Mathematically, this frameworks is defined as Markov Decision Process (MDP) given by $\mathcal{M}:=\{\mathcal{S}, \, \mathcal{A}, \, \mathbb{P},\,r,\, \gamma \}$, where $\gamma\in(0,1)$ is the discount factor which decides the importance of future rewards for each instant.  The state space $\mathcal{S} \subseteq \mathbb{R}^{q}$ and actions space $\mathcal{A} \subseteq  \mathbb{R}^{p}$ is continuous. Hence, we hypothesize that the agent selects actions $a_t\sim \pi(\cdot |s)$ over a time invariant distribution denoted by $\pi(\cdot |s)$ for a given state $s$. The distribution $\pi(\cdot |s)$ is called a policy which controls the probability of taking a particular action $a$ in given state $s$.  The goal in the RL problem is to search for policy $\pi(\cdot |s)$ such that the average cumulative reward return  (called value) is maximized given by :
%%%%%%%%%%%%%%%%%%%%%%%%%%%%%%%%%%%%%%%%%%%%%%%%%%%%%%%%%%%%%%%%%%
%%%%%%%%%%%%%%%%%%%%%%%%%%%%%%%%%%%%%%%%%%%%%%%%%%%%%%%%%%%%%%%%%%%%%%%%%%%%%%%%
\begin{align}\label{eq:value_func}
V^{\pi}(s) = \mathbb{E} \bigg[ \sum_{t=0}^{\infty} \gamma^{t} r(s_t,a_t)~|~s_0=s, a_t\sim \pi(\cdot|s_t)  \bigg], 
\end{align}
%
%\begin{figure}[t]
%	%\centering
%	%\vspace{4mm}
%	{\includegraphics[scale=0.15]{figs/new.png}} 	\includegraphics[scale=0.05]{figs/mario.jpg}
%	\caption{\scriptsize 1D Mario environment \cite{matheron2019problem}.}
%	\vspace{0mm}   
%			\caption{ } \label{fig:env}\vspace{-0mm} 
%\end{figure}
 
where $V^{\pi}(s)$ is the value function with respect to state $s$, and $s_0$ denotes the initial state along a trajectory $\{s_t,a_t,r(s_t,a_t)\}_{u=0}^\infty$. Similar to fixing the initial state $s_0$, if we fix initial action as well $a_0=a$, the we can write the action-value function as 
\begin{align}\label{eq:Q_value}
	Q^{\pi}(s,a)=\mathbb{E} \bigg[ \sum_{t=0}^{\infty} \gamma^{t} r(s_t,a_t)~|~s_0=s,a_0=a, a_t\sim \pi(\cdot|s_t)  \bigg].
\end{align}
We note that the expectation in \eqref{eq:value_func}-\eqref{eq:Q_value} is with respect to the product measure of policy $a_t \sim \pi(\cdot |s_t)$ and state transition density $s_{t+1} \sim \mathbb{P}(.|s_t,a_t)$. The selection of action $a_t$ would control the possibility of visiting different state in the state space $\mathcal{S}$, and hence also responsible for exploring the state space. This also becomes important because in this work, we are specifically interested in environments where the rewards are sparse (cf. Sec.~\ref{sec:experiments}). By sparse rewards we mean that they are available once in a while (see Fig. \ref{fig:b}) or there are high reward states available (see Fig. \ref{fig:a}) but too far in the state space. Learning a good policy in such environments is a difficult task and that is the focus of this work.
Hence, the goal here is to find a policy $\pi$ such that
\begin{align}\label{main}
	\max_{\pi} V(s_0),
\end{align}
with $s_0\sim\rho_0(s)$ and $\rho_0(s)$ being an arbitrary initial state distribution. Since, $\pi$ here is a policy distribution, it becomes intractable to solve the problem  \eqref{main} in it general form and we keep our focus to a search over parameterized class of policies denoted by $ \pi_{\bm{\theta}}(\cdot |s_t)$ where $\bbtheta$ is the parameter which defined the policy distribution completely. So now, our search over distributions boil down to search over set of parameters $\bbtheta$ \cite{sutton2018reinforcement} given by  
\begin{align}\label{eq:main_prob}
\max_{\bm{\theta}} J(\bm{\theta}) : = V^{{\pi}_{\bm{\theta}}}(s_0),
\end{align}
with $s_0\sim\rho_0(s)$. We note that the problem in \eqref{eq:main_prob} is non-convex with respect to optimization variable $\bbtheta$. Next, we derive the standard policy gradient algorithm to solve the problem in \eqref{eq:main_prob} and discuss challenges in the sparse reward settings. 

%where, objective is given by $J(\bm{\theta}) $. Observe that \eqref{eq:main_prob} is non-convex in $\bm{\theta}$, and therefore, finding the optimal policy is challenging even in the deterministic setting. However, in RL, the search procedure necessarily interacts with the transition dynamics $\mathbb{P}(s'|s,a)$ as well. Before detailing how one may implement first-order stochastic search to solve \eqref{eq:main_prob}, we introduce the widely used standard Gaussian policy parameterization, and clarify how its practice can lead to hidden bias.

%%%%%%%%%%%%%%%%%%%%%%%%%%%%%%%%%%%%%%%%%%%%%%%%%%%%%%%%%%%%%%%%%%%%%%%%%%%%%%%%%%%%%%%%%%%%%%%%%%%%%%%%%%%%%%%%%%%%%%%%%%%%%%%%%%%%%%%%%%%%%%%%%%%%%%
%\subsection{Example Policy Parameterizations}\label{subsec:examples}
%

%%%%%%%%%%%%%%%%%%%%%%%%%%%%%%%%%%%%%%%%%%%%%%%%%%%%%%%%%%%%%%%%%%%%%%%%%%%%%%%%%%%%%%%%%%%%%%%%%%%%%%%%%%%%%%%%%%%%%%%%%%%%%%%%%%%%%%%%%%%%%%%%%%%%%%%%%%%%%%%%
\subsection{Policy Gradient Algorithm}\label{section:algorithm}
The policy gradient (PG) algorithm is a well known technique to perform search for optimal parameters $\bbtheta$ in parameter space $\bbtheta \in \mathbb{R}^d$. The key result which enables us to write policy gradient for the complicated objective in  \eqref{eq:main_prob} is the Policy Gradient Theorem \cite{sutton2018reinforcement}, which states that the gradient of $J(\bbtheta)$ with respect to $\bbtheta$ can be written as 
\begin{align}
	\nabla J(\theta)&=\int_{\mathcal{S}\times \mathcal{A}}\sum_{t=0}^\infty\gamma^t \cdot \mathbb{P}(s_k=s\given s_0,\pi_\theta)\times 
	\nonumber
	\\
&\hspace{2cm}	\times \nabla \pi_{\theta}(a\given s)\cdot Q_{\pi_\theta}(s,a)\cdot dsda\label{equ:policy_grad_1}
\\
	&=\frac{1}{1-\gamma}\int_{\mathcal{S}\times \mathcal{A}}(1-\gamma)\sum_{t=0}^\infty\gamma^t \cdot \mathbb{P}(s_k=s\given s_0,\pi_\theta)\times 
	\nonumber
	\\
	&\hspace{2cm}\times \nabla \pi_{\theta}(a\given s)\cdot Q_{\pi_\theta}(s,a)\cdot dsda\notag
	\\
	&=\frac{1}{1-\gamma}\int_{\mathcal{S}\times \mathcal{A}}\rho_{\pi_\theta}(s) \cdot \pi_{\theta}(a\given s)\times 
\nonumber
\\
&\hspace{2cm} \times \nabla \log[\pi_{\theta}(a\given s)]\cdot Q_{\pi_\theta}(s,a)\cdot dsda\notag\\
	&=\frac{1}{1-\gamma}\cdot\EE\big[\nabla\log\pi_{\theta}(a\given s)\cdot Q^{\pi_\theta}(s,a)\big], \label{eq:policy_grad}
\end{align}
and the expectation in \eqref{eq:policy_grad} is over $(s,a)\sim \rho_{\theta}(\cdot,\cdot)$ where   $\rho_{\theta}(s,a)$$=$$\rho_{\pi_\theta}(s)\cdot \pi_{\theta}(a\given s)$ now denotes a valid probability distribution function also called as  \emph{discounted state-action  occupancy measure} over continuous state and action spaces. From the expressions of $\rho_{\theta}(s,a)$ note that the selection of policy class has a significant affect on the eventually occupancy measure induced. In tabular MDP settings, to make sure the convergence to global optimal, an assumption of \emph{persistent exploration} is needed \cite{mei2020global}, which is satisfied by making sure that $\pi(a\given s)>0$ for all $s$ and $a$. We remark that satisfying such assumption automatically takes care of the fact that we explore almost all parts of the state space because the probability of reaching any other state $s'$ is not zero because of $\pi(a\given s)>0$. Therefore, in tabular MDP, things work well even in the sparse reward settings.  In contrast, in continuous action spaces, imposing such assumption $\pi(a\given s)>0$ on the policy distribution $\pi(a\given s)>0$ would violate the integrable assumption of probability distributions, and hence is not a valid assumption. So the induced exploration in the state space is mostly controlled by the policy distribution class we choose for parametrization. The standard parametrization class which is widely used in the literature is Gaussian \cite{kang2018policy,rengarajan2022reinforcement,zhang2020global,yuan2020stochastic}, and given as follows. 
%

%%%%%%%%%%%%%%%%%%%%%%%%%%%%%%%%%%%%%%%%%%%%%%%%%%%%%%%%%%%%%%%%%%%%%%%%%%%%%%%%%%%%%%%%%%%%%%%%%%%%%%%%
\begin{example}[Gaussian Parametrization]\normalfont \label{eg:gaussian_fixed_var}
	We assume that the policy $\pi_{\bbtheta}(a\given s)$ is a Gaussian distribution given by 
\begin{align}\label{eq:Gaussian_policy}
\pi_{\bm{\theta}}(a|s) = \mathcal{N} (a|\varphi(s)^\top \bm{\theta}, \sigma^2),
\end{align}
	where $\bbtheta $ controls the mean of the Gaussian, $\varphi(s)$ denotes the states feature representation $\varphi: \mathcal{S}\rightarrow \mathbb{R}^d$ with $d\ll q$, and $\sigma^2$ is fixed variance.  We can make $\sigma$ as a parameter as well we avoid that for the sake of explanation simplicity. 
\end{example}
Now, specifically for sparse reward settings, one major drawback of Gaussian parametrization for policy is its tendency to take actions close to mean value. This feature would restrict the model transition to a state $s'$ which is farther from current state $s$ due to action selection $a\sim \mathcal{N} (\varphi(s)^\top \bm{\theta}, \sigma^2)$ close to mean value. This induces a limited exploration for the algorithm, and it fails to learn in sparse reward environments. To deal with this issue, different techniques such as information maximization \cite{reward_2} and learning from demonstrations \cite{kang2018policy} are proposed. But the main disadvantages of such techniques are that entropy regularization required the estimation of the density function of occupancy which is quite expensive, and prior demonstrations could be quite bad and lead to completely irrelevant policies. Hence, to deal with such issues, instead of proposing any augmentation to existing techniques to handle sparse rewards, we resort to a completely novel approach and proposed to utilize heavy-tailed distributions to parameterize the policy $\pi_{\bbtheta}$. We explain this idea in detail in the next section. 

%is a probability distribution that denotes the \emph{discounted state-action  occupancy measure}, which is the product of the discounted state occupancy measure $\rho_{\pi_\theta}(s)$$=$$(1-\gamma)\sum_{t=0}^\infty\gamma^t \mathbb{P}(s_k=s\given s_0,\pi_\theta)$ and  policy $\pi_{\theta}(a\given s)$. In \cite{sutton2000policy}, both $\rho_{\pi_\theta}(s)$ and $\rho_{\bm{\theta}}(s,a)$ are established as valid probability distributions.
%
\section{Proposed Heavy-Tailed Stochastic Policy Gradient for Sparse Rewards}\label{sec:proposed}
In this section, we present the main idea of this work and develop a stable heavy-tailed stochastic policy gradient descent algorithm to deal with sparse reward settings. 

\subsection{Heavy-Tailed Policy Parametrization}  As a first step towards developing such an algorithm, we propose to parameterize the policy by a class of heavy-tailed distributions. An example of heavy such parametrization is Cauchy distribution which is given by
%which we provide next.  
%
%	\begin{example}[Cauchy Parametrization]\normalfont \label{eg:alpha_stable}
	%
	%\textcolor{blue}{please present formula for a general alpha stable policy and then Cauchy policy in this example. Include references}
	%%%%%%%%%%%%%%%%%%%%%%%%%%%%%%%%%%%%%%%%%%%%%%%%%%%%%%%%%%%%%%%%%%%%%%%%%%%%%%
% A generalized version of  centered  Gaussian distributions is called symmetric $\alpha$ stable $\mathcal{S}\alpha\mathcal{S}$ distribution with $\alpha \in  (0, 2]$  as the tail index  which specifies the heaviness of the tails. \cite{nguyen2019first}. For such distributions, we denote random variable $\textbf{X} \sim \mathcal{S}\alpha\mathcal{S} (\sigma)$ which is associated with characteristic function $\mathbb{E}\left[ e^{i\omega \textbf{X}}\right]  = e^{-| \sigma|\omega^{\alpha}}$ and scale parameter $\sigma  \in (0, \infty) $. For $\alpha=1$, it boils down to Cauchy distribution given by 
	%
	\begin{align}\label{eq:Cauchy_policy}
		\pi_{\bm{\theta}}(a|s) = \frac{1}{\sigma \pi (1+((a-\varphi(s)^\top \bm{\theta})/\sigma)^2) },
	\end{align}
	%%%%%%%%%%%%%%%%%%%%%%%%%%%%%%%%%%%%%%%%%%%%%%%%%%%%%%%%%%%%%%%%%%
	%		%%%%%%%%%%%%%%%%%%%%%%%%%%%%%%%%%%%%%%%%%%%%%%%%%%%%%%%%%%%%%%%%%%%%%%%%%%%%%%%%
			where $\sigma$ is the fixed scaling parameter.  
			%For non-integer (fractional) value of  $\alpha$, the distribution does not exhibit a closed form expression, and is referred to as fractal \cite{mandelbrot1982fractal}. 
%			
%\end{example} 
%
Other heavy-tailed distributions include the Extreme value distribution, Weibull distribution, log-normal distribution, Student's t distribution, Generalized Gaussian distribution, etc. The Laplace distribution has also fatter tails than the Gaussian distribution. In the financial literature, such distributions have been associated with the phenomenon of "black swan" events \cite{taleb2007black,taylor2009black}.

With the policy parametrization specified, next goal is to compute the policy gradient mentioned in \eqref{eq:policy_grad}. But the challenge is the transition model dynamics are assumed to me unknown so it is not possible to evaluate $	\nabla J(\theta)$ in closed form. So we take stochastic approximation approach and evaluate the stochastic gradient estimate. To write that, consider a randomized horizon $T_k\sim\text{Geom}(1-\gamma^{1/2})$ with trajectory sample  $\{(s_0,a_0)\cdots(s_{T_{k}},a_{T_{k}})\}=:\xi_k(\bbtheta_k)$, then stochastic gradient can be written as 
\begin{align}\label{eq:policy_gradient_iteration}
	{\nabla}  J(\bm{\theta}_k,&\xi_k(\bbtheta_k))\\
	=& \sum_{t=0}^{T_k}\gamma^{t/2}r(s_t,a_t)\cdot\bigg(\sum_{\tau=0}^{t}\nabla\log\pi_{\bm{\theta}_k}(a_{\tau}\given s_{\tau})\bigg),\nonumber
\end{align}
where $	{\nabla}  J(\bm{\theta}_k,\xi_k(\bbtheta_k))$ denotes the unbiased estimator of gradient $	\nabla J(\theta_k)$ at $\bbtheta_k$ (see \cite[Lemma 1]{bedi2021sample} for proofs) and 
  and $\xi_k(\bbtheta_k)$ denotes the randomness in the estimate at $k$. Note the variable horizon length of the trajectories in \eqref{eq:policy_gradient_iteration} which is important to obtain an unbiased estimator. Otherwise, a fixed horizon length estimators where $T_k=H$ for all $k$ (as in  \cite{papini2018stochastic,yuan2020stochastic}), results in  a bias-variance tradeoff for gradient estimate \cite{baxter2001infinite}. Further, note the summation over two indexes in  \eqref{eq:policy_gradient_iteration} $t$ corresponds to the rollout trajectory, and $\tau$ collects score function till $t$ from the starting. With the stochastic gradient defined in \eqref{eq:policy_gradient_iteration}, the heavy tailed stochastic policy gradient iterate is given by
\begin{align}\label{eq:policy_gradient_iteration22}
	\bm{\theta}_{k+1} =& \bm{\theta}_k + \eta {\nabla} J(\bm{\theta}_k, \xi_k(\bbtheta_k)), \; 
\end{align}
where  $\eta>0$ denotes the step size.  %The overall procedure is summarized in Algorithm \ref{Algorithm1}. 
%
%%%%%%%%%%%%%%%%%%%%%%%%%%%%%%%%%%%%%%%%%%%%%%%%%%%%%%%%%%%%%%%%%%
%%%%%%%%%%%%%%%%%%%%%%%%%%%%%%%%%%%%%%%%%%%%%%%%%%%%%%%%%%%%%%%%%%%%%%%%%%%%%%%%
%
Note that the score function to evaluate the stochastic policy gradient in \eqref{eq:policy_gradient_iteration22} is parameterized by a heavy-tailed policy due to the sparse rewards settings considered in this work. While this selection of heavy tailed parametrization serves the purpose of selecting actions far from mean and induce sufficient exploration into the algorithm behavior, this exhibits a downside as well. The resulting algorithm tends to be unstable to to heavy tails and probability of taking extreme actions. We mitigate this issue by introducing a momentum based gradient tracking to the proposed algorithm which is the focus of next subsection.

\subsection{Stable Heavy-Tailed Stochastic Policy Gradient Algorithm}
The direct replacement of Gaussian policy parametrization with heavy-tailed policy parametrization results in an unstable behavior for the algorithm because the stochastic gradient estimates exhibit high variations from one sample to other. To deal with this issue, we need to invoke the idea of introducing momentum to stochastic gradient (SG) updates which has been successfully used in other machine learning approaches \cite{kingma2014adam}. Hence, we replace the update in \eqref{eq:policy_gradient_iteration22} as follows
\begin{align}
		\bm{g}_{k} =& (1-\beta)\bm{g}_{k-1} + \beta {\nabla} J(\bm{\theta}_k, \xi_k(\bbtheta_k)), \label{momentum} \; \\
	\bm{\theta}_{k+1} =& \bm{\theta}_k + \eta 	\bm{g}_{k},\label{eq:policy_gradient_iteration222} \; 
\end{align}
where $\beta$ is the tuning parameter and update in \eqref{momentum} is called the momentum update. Note that for a small $\beta$ (say $\beta=0.2$) would results in utilizing the exponential average of past gradients rather than just considering the current stochastic gradient ${\nabla} J(\bm{\theta}_k, \xi_k(\bbtheta_k))$. This update is popular in the SG descent literature and achieves significant improvement empirically as compared to special case of $\beta=1$ [13 from \cite{cutkosky2019momentum}] but does not result in theoretical gain. To address this issue, the authors in \cite{cutkosky2019momentum} have proposed a modified momentum based gradient tracking which result in provable variance reduction. With motivation from results in \cite{cutkosky2019momentum}, we propose a novel gradient tracking scheme presented next for stochastic policy gradients with heavy-tailed policy parametrization as
\begin{algorithm}[t]
	\begin{algorithmic}[1]
		%\textbf{Procedure:}{Require: Step-size $\alpha_k$}
		\STATE \textbf{Initialize} :   Initial parameter $\bm{\theta}_0$, momentum parameter $\beta$, discount factor $\gamma$, step-size $\eta$, and gradient estimate $\bbg_0$$ =$$0$  \\
		\textbf{Repeat for $k=1,\dots$}
		\STATE Sample two trajectories $\xi_k(\bbtheta_{k})$ and $\xi_k(\bbtheta_{k-1})$ of length  $T_k\sim\text{Geom}(1-\gamma^{1/2})$ using policies $\pi_{\bm{\theta}_{k}}$ and $\pi_{\bm{\theta}_{k-1}}$, respectively%=\{(s_0,a_0)\cdots(s_{T_{k+1}},a_{T_{k+1}})\}$ where 
		\vspace{-0mm}
		%\STATE   Estimate $\hat{Q}^{{\pi}_{\bm{\theta}_{k}} } (s_{T_{k+1}},a_{T_{k+1}})$ via Monte-Carlo rollout \eqref{eq:rollout}
		\STATE  Estimate ${\nabla} J(\bm{\theta}_k, \xi_k(\bbtheta_{k}))$, ${\nabla} {J} (\bm{\theta}_{k-1}, \xi_k(\bbtheta_{k-1}))$  via \eqref{eq:policy_gradient_iteration} and \eqref{eq:policy_gradient_iteration22222}, respectively \vspace{-0mm}
		\STATE  Estimate $\bbg_k$ via \eqref{momentum2} \vspace{-0mm}
		\STATE Update $	\bm{\theta}_{k+1} = \bm{\theta}_k + \eta 	\bm{g}_{k}$
		\STATE $k \leftarrow k+1$
		
		\textbf{ Until Convergence}
		% \EndProcedure
		\STATE \textbf{Return: $\bm{\theta}_k$} \\ 
	\end{algorithmic}
	\caption{Heavy-Tailed Stochastic Policy Gradient (HTSPG)}
	\label{algo:one}
\end{algorithm}
\begin{align}
	\bm{g}_{k} =& (1-\beta)\bm{g}_{k-1} + \beta {\nabla} J(\bm{\theta}_k, \xi_k(\bbtheta_k)) 
\label{momentum2}
	\\
	& +(1-\beta) ({\nabla} J(\bm{\theta}_k, \xi_k(\bbtheta_k))-{\nabla} J(\bm{\theta}_{k-1}, \xi_k(\bbtheta_{k-1}))	), \nonumber \; \\
	\bm{\theta}_{k+1} =& \bm{\theta}_k + \eta 	\bm{g}_{k},\label{eq:policy_gradient_iteration222} \; 
\end{align}
 where ${\nabla} J(\bm{\theta}_{k-1}, \xi_k(\bbtheta_{k-1}))$ denotes the another stochastic gradient evaluated at instant $k$ with policy parameter $\theta_{k-1}$. The explicit expression is given by 
\begin{align}\label{eq:policy_gradient_iteration22222}
	{\nabla} J(\bm{\theta}_{k-1}, &\xi_k(\bbtheta_{k-1}))\\
	=& \sum_{t=0}^{T_k}\gamma^{t/2}r(s_t',a_t')\cdot\bigg(\sum_{\tau=0}^{t}\nabla\log\pi_{\bm{\theta}_{k-1}}(a_{\tau}'\given s_{\tau}')\bigg),\nonumber
\end{align}
where $\xi_k(\bbtheta_{k-1}):=\{s_i',a_i',r(s_i',a_i')\}_{i=0}^{T_k}$ denotes the trajectory generated bu using policy parameter $\bbtheta_{k-1}$ but at instance $k$. 
Note that there will be two Monte Carlo trajectories required to perform the update in \eqref{momentum2}. We remark an important difference of update in \eqref{momentum2} to the gradient tracking proposed in \cite{cutkosky2019momentum}.  The momentum step in \cite[Eq. (2)]{cutkosky2019momentum} would require the use of ${\nabla} J(\bm{\theta}_{k-1}, \xi_k(\bbtheta_{k}))$ (to keep the stochastic quantity same) instead of ${\nabla} J(\bm{\theta}_{k-1}, \xi_k(\bbtheta_{k-1}))$ which we propose to use in this work. The use of term ${\nabla} J(\bm{\theta}_{k-1}, \xi_k(\bbtheta_{k}))$ has been proposed in the literature for reinforcement learning settings in \cite{yuan2020stochastic} along with importance sampling weight adjustments to take care of the distributional shift which occurs due to the dependence of stochastic trajectory $\xi_k(\bbtheta_{k})$ on $\bbtheta_k$.  Next, we intuitively explain why it makes sense to use the update in \eqref{momentum2} and it helps to reduce the variance of stochastic gradients, and hence results in a stable algorithm.

 To understand it, let us consider the stochastic error introduced to the original gradient due to \eqref{momentum2} as $\epsilon_k=\bm{g}_{k}-{\nabla} J(\bm{\theta}_k)$. We note that $\epsilon_k$ defines the stochastic error in the gradient direction to perform the ascent update, and if we show that $\mathbb{E}\|\epsilon_k\|^2$ has a decreasing behavior with respect to $k$, this implies that the proposed momentum based update has resulted in variance reduction. Let us look at the explicit expression of $\epsilon_k$ as 
 \begin{align}
\epsilon_k =&(1-\beta)\epsilon_{k-1} + \beta ( {\nabla} J(\bm{\theta}_k, \xi_k(\bbtheta_k))-{\nabla} J(\bm{\theta}_k)) \nonumber
\\
& +(1-\beta) ({\nabla} J(\bm{\theta}_k, \xi_k(\bbtheta_k))-{\nabla} J(\bm{\theta}_{k-1}, \xi_k(\bbtheta_{k-1})))	\nonumber\\
& \qquad\qquad\qquad+ (1-\beta)({\nabla} J(\bm{\theta}_k)-{\nabla} J(\bm{\theta}_{k-1})).\label{momentum222}
 \end{align}

Next, note that it is the second, third, and fourth term on the right hand side of \eqref{momentum222} which we need to control. We can easily control the second term on the right hand side of \eqref{momentum222} by keeping $\beta$ small. From the smoothness of $J$, we know that $ \|{\nabla} J(\bm{\theta}_k)-{\nabla} J(\bm{\theta}_{k-1})\|\approx \mathcal{O}\left(\eta\|\bbtheta_{k}-\bbtheta_{k-1}\|\right)$ which can be controlled by step size $\eta$. The only  remaining term is $\|{\nabla} J(\bm{\theta}_k, \xi_k(\bbtheta_k))-{\nabla} J(\bm{\theta}_{k-1}, \xi_k(\bbtheta_{k-1}))\|$ which can also be assumed $\approx \mathcal{O}\left(\eta\|\bbtheta_{k}-\bbtheta_{k-1}\|\right)$ when $\bbtheta_k$ and $\bbtheta_{k-1}$ are close to each other. This is possible because of the dependence of trajectories $\xi_k(\cdot)$ on $\bbtheta$ which is not the case in \cite{cutkosky2019momentum}. Therefore, by controlling $\beta$ and $\eta$, it is possible to develop a stable algorithm with heavy-tailed policy parametrizations. We summarize the algorithm steps in Algorithm \ref{algo:one}.  Further, we extensively test the empirical performance of the purposed algorithm on different sparse environments in next section and show the performance benefits achieved in practice. We defer the theoretical analysis of the proposed algorithm to future scope of this work. 
\begin{figure}[t]
	\centering
	\includegraphics[scale=0.4]{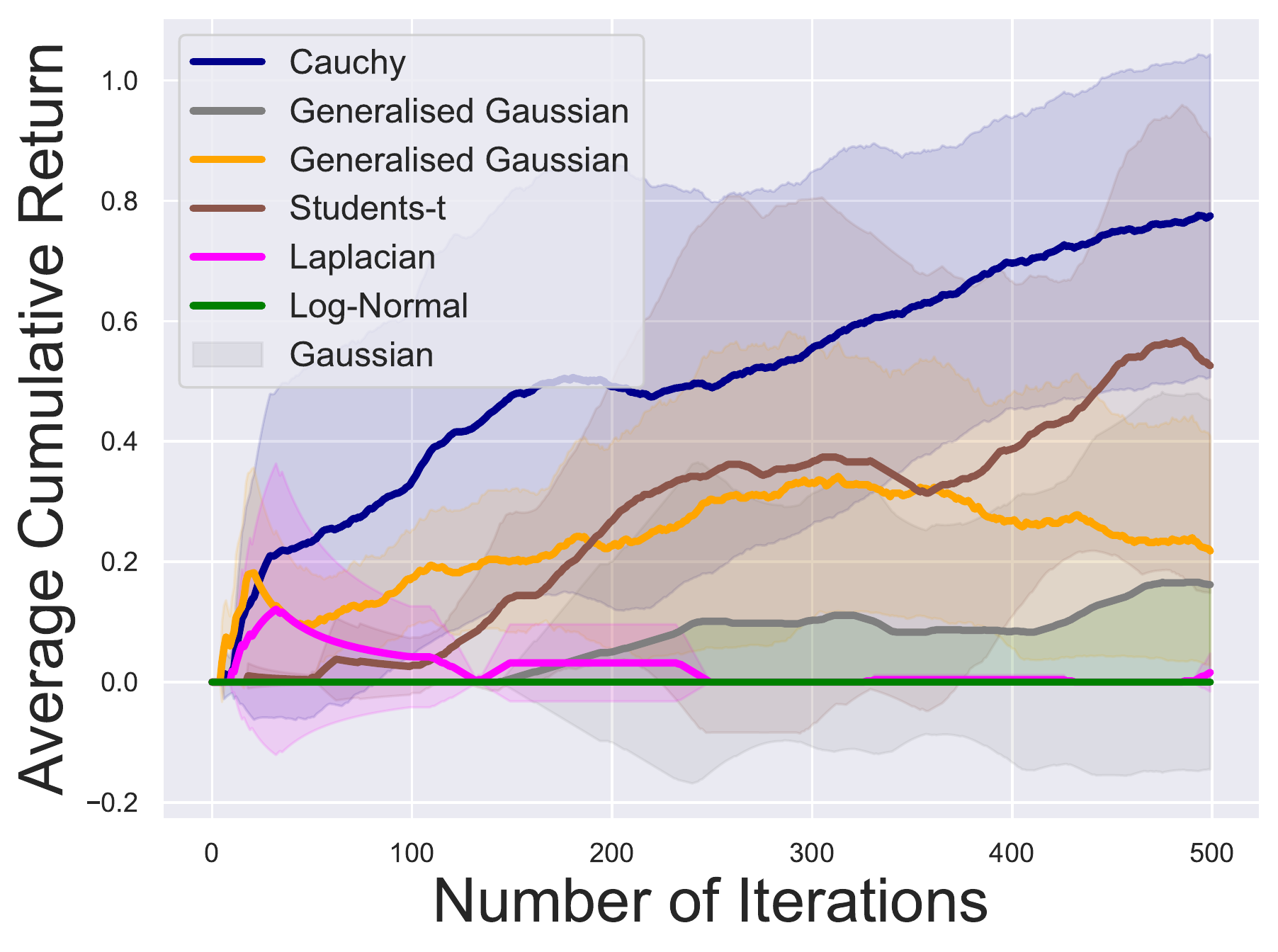}
	\caption{\normalsize In this figure, we show the importance of selecting Cauchy as our heavy-tailed policy as compared to other possible policy parametrization. We run tests on 1D Mario continuous control environment and plot the average reward return for different policy parametrizations. It is clear that Cauchy performs the best among all of them and achieves the highest reward return. We specifically demonstrate the superiority of Cauchy's performance over other heavy-tailed distribution both in terms of rewards and improved speed of convergence}
	\label{fig:starting}\end{figure}

\section{Experiments}\label{sec:experiments}
\begin{figure*}[t]
	\centering
	\subfigure[1D Mario Environment \cite{matheron2019problem}.]{\label{fig:mario}		\includegraphics[width=0.3\textwidth]{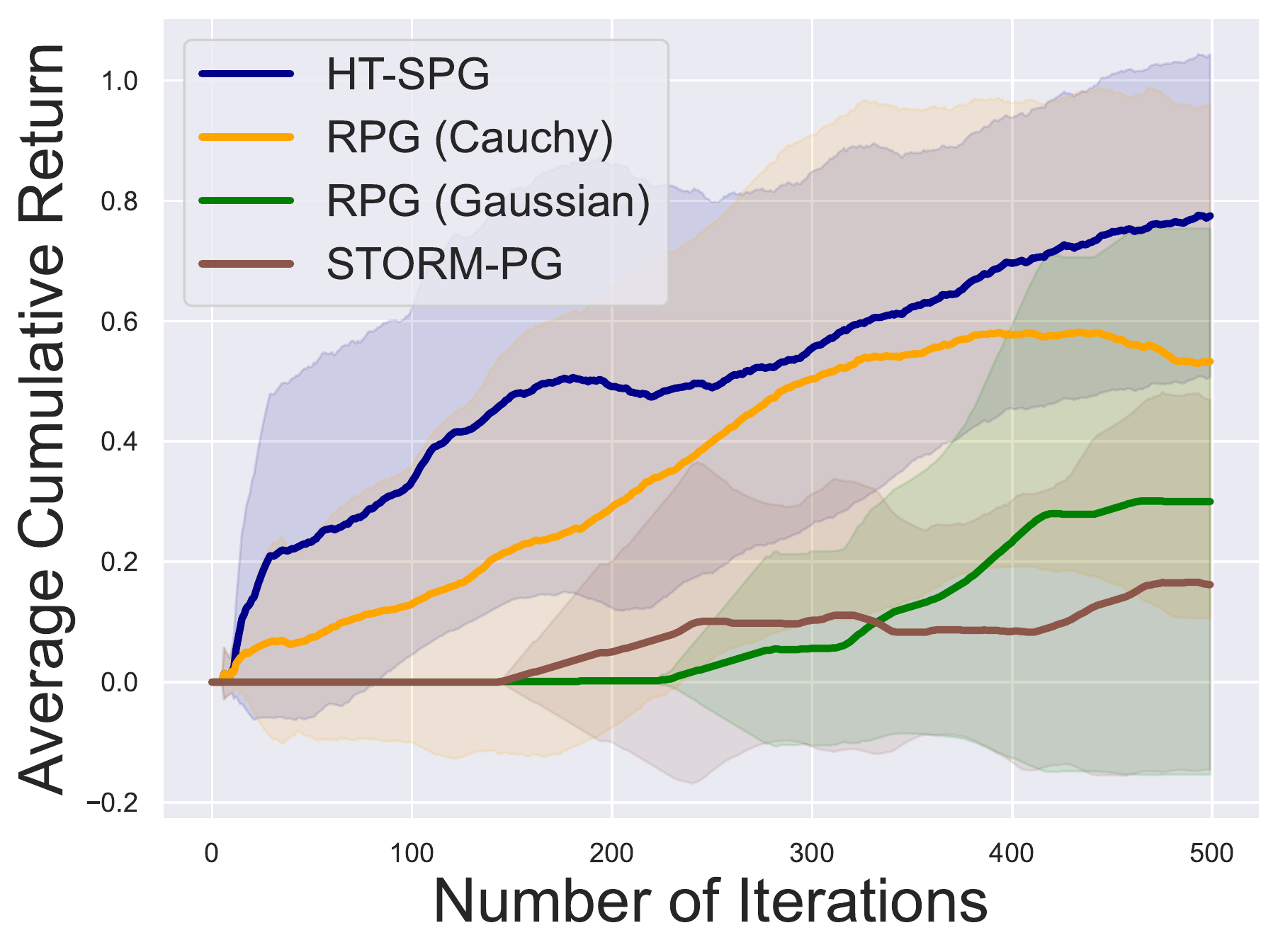}}
	\subfigure[Pathological Mountain Car.]{\label{fig:PMC}		\includegraphics[width=0.3\textwidth]{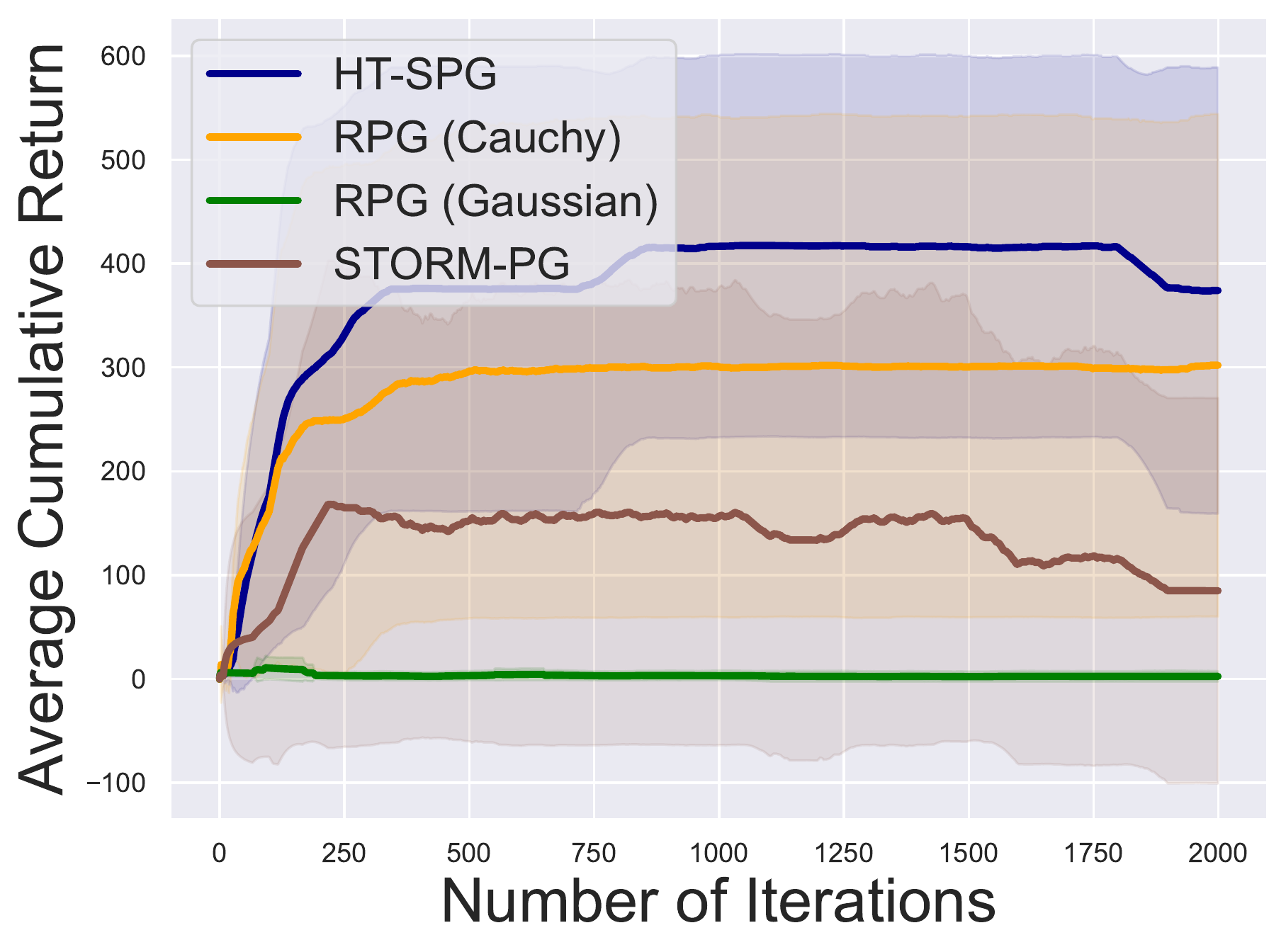}}
		\subfigure[Sparse Pendulum.]{\label{fig:sparse_pendulum}		\includegraphics[width=0.3\textwidth]{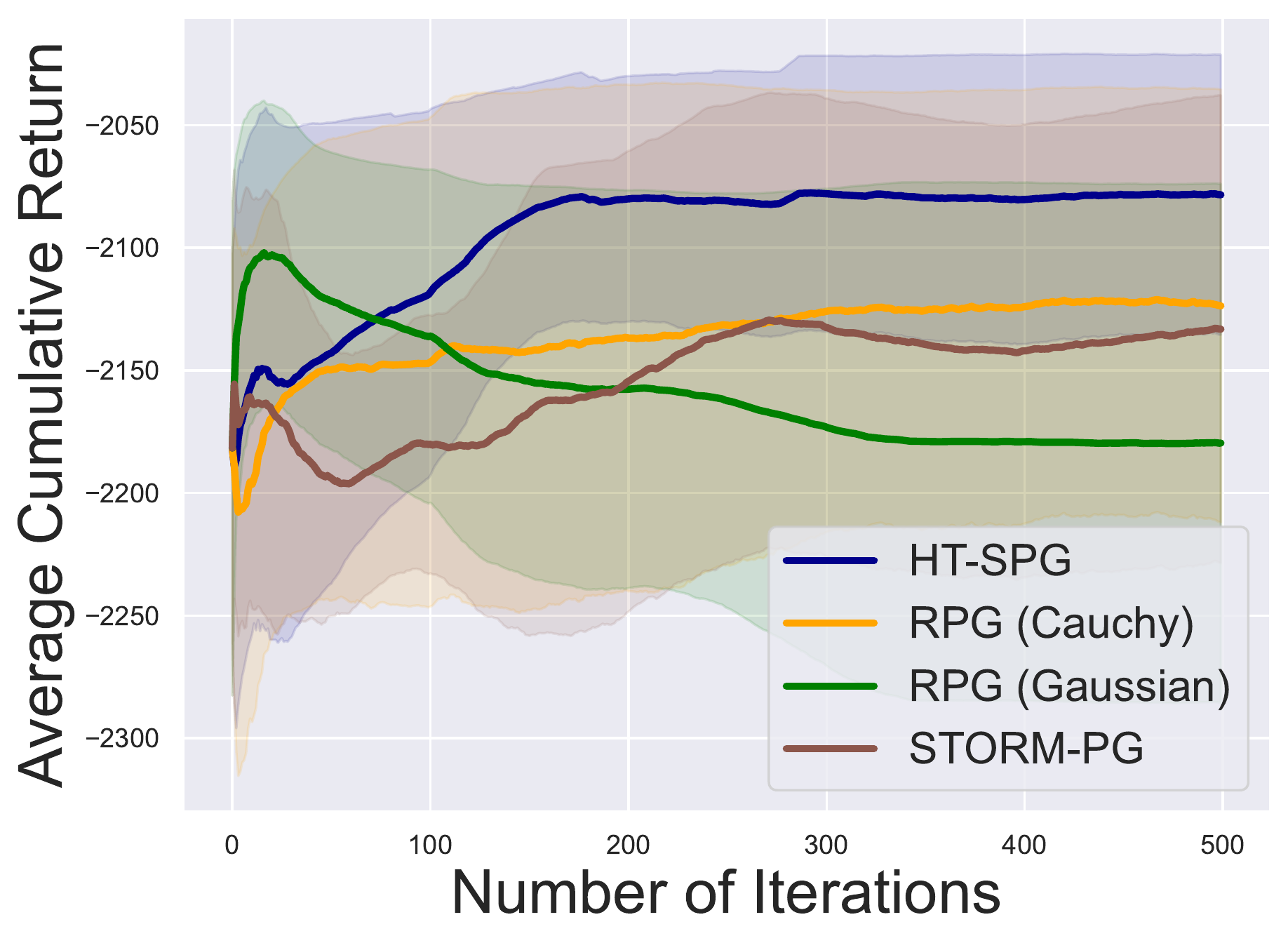}}
	\caption{\normalsize In this figure, we compare the performance of the proposed HT-SPG algorithm with RPG \cite{zhang2020global} and STORM-PG \cite{yuan2020stochastic} which are state-of-the-art algorithms to solve the same problems without expert's demonstrations. Here, RPG (Cauchy) denotes the RPG algorithm with Cauchy policy parametrization and we compared to it to show that just replacing Gaussian with Cauchy is not the best thing to do. It works but results in high variance in the reward returns as shown by the high confidence intervals of yellow line. We plot the average cumulative reward return with respect to number of iterations/episodes for (a) 1D Mario environment \cite{matheron2019problem}, (b) Pathological Mountain Car (cf. \ref{fig:a}), and (c) Sparse Pendulum of OpenAI Gym environments. We note that the HT-SPG is able to achieve highest reward return consistently in all the environments.Note : Total number of training samples = Batch size * No of Iterations.}
		\label{proof_of_concept}
	\end{figure*}
In this section, we proceed to perform extensive experimental validation of the proposed ideas in this work. 
First, we perform a detailed analysis and performance comparison of proposed stable Heavy-Tailed Stochastic Policy Gradient Algorithm (HT-SPG) in classic continuous control reinforcement learning environments with sparse and complex rewards such as 1D Mario \cite{matheron2019problem}, Pathological Mountain Car (cf. Fig. \ref{fig:a}), and Sparse Pendulum \cite{brockman2016openai}. 
Second, to test the performance on complicated continuous environments, we consider the Sparse MuJoCo environments namely Hopper-v2 as done in \cite{rengarajan2022reinforcement}. Finally, we compare the performance of HT-SPG against state-of-the-art LOGO algorithm \cite{rengarajan2022reinforcement} and show consistent performance improvements under complex and challenging settings.

\textbf{Importance of Policy Parametrization:} Before discussing the main experimental results, we start by demonstrating (see Fig.~\ref{fig:starting}) the limitations of light-tail policy parametrization and emphasize the importance of using heavy-tail distributions such as Cauchy for policy parametrization. Fig.~\ref{fig:starting} shows the average cumulative reward return for different policy parametrizations in a 1D Mario environment. We demonstrate that Cauchy distribution-based policy is able to achieve the highest reward return in the most sample-efficient manner. This is mainly due to the better exploratory behavior achieved by the Cauchy-based policy as compared to other policies. Hence, we will be using Cauchy policy parametrizations for the rest of the experiments. We detail the different environment settings as follows. 
  
\subsection{Learning Without Demonstrations}
In this subsection, we run experiments in sparse reward environments and compare against other state-of-the-art algorithms which operate without any access to expert's demonstrations. The details of environments are as follows. 

\begin{itemize}
	\item  \textbf{1D Mario Environment}: This is a  one-dimensional, discrete-time, continuous state and action space environment (cf. Fig.\ref{fig:a}). The state space is $s\in[0,1]$ and action space is $a\in[-0.1,0.1]$. The goal is to collect the coin place at $s=0$ and agent can move in right or left by any amount between $[-0.1,0.1]$. The reward is defined as
    $r(s_t,a_t)=\mathbbm{1}_{\{s_t+a_t <0\}}$ and transition model as $s_{t+1}=\min\{1,\max{\{0,s_t+a_t\}}\}$. We note that reward is sparse because it is 1 only at the goal, otherwise it's zero in the full state space. Each of the episodes are initialized at $s_0=0.9$.  
	
	\item \textbf{Pathological Mountain Car}: This is a continuous state action space environment with misaligned goal (see Fig. \ref{fig:a}). The reward is distributed widely over the state space with a low reward state and a bonanza top a higher hill.   The low reward state is at $s = 2.667$ with a reward of $10$ and a high reward state farther apart at $s=-4.0$ with $500$ units of reward. For PMC, we consider a reward structure in which the amount of energy expenditure, i.e., the action squared, at each time-step is negatively penalized, as given by 
    \begin{align} 
    \label{equ:reward_mc}
     r(s,a) = -  &a_t^2\mathbbm{1}_{\{-4.0 < s < 3.709, s \neq 2.667\}} 
     \nonumber
     \\
     &+(500 -  a_t^2)\mathbbm{1}_{\{s = -4.0\}}
     \nonumber
     \\
     &\quad+ (10-   a_t^2)\mathbbm{1}_{\{s = 2.67\}}.
    \end{align}
    Here , the action is denoted as $a$ and is a one-dimensional scalar which represents the speed of the vehicle $\dot{s}_t$.

	\item \textbf{Sparse Inverted Pendulum}: This is an unstable inverted pendulum (pole) attached to a cart (see Fig.~\ref{Hopper}), and the goal is to keep the pole upright \cite{brockman2016openai}. An agent can move the cart to the left or right via applying a discrete force of $\pm 1$ along the horizontal axis of the cart. It is exactly like Open AI gym's Pendulum-v0, but with sparse rewards.
	
\end{itemize}

We run the proposed algorithm HT-SPG in the above-mentioned environments and compare with other state-of-the-art existing algorithms with light-tailed policy parametrization (Gaussian) such as RPG \cite{zhang2020global}, and STORM-PG \cite{yuan2020stochastic}. There is a state-of-the-art algorithm to solve continuous control problems without any demonstrations. We present the results in Fig. \ref{proof_of_concept}, where RPG (Cauchy) denotes RPG algorithm with Cauchy policy parametrization. It is included to show that just replacing Gaussian (RPG (Gaussian)) with Cauchy is not sufficient to achieve the desired performance, and it results in unstable behavior which exhibits high variance in the reward returns. This issue is corrected by using the momentum-based tracking in HT-SPG.  In all these classic continuous control environments with sparse rewards, our proposed HT-SPG algorithm outperforms all the other methods based on light-tail distribution, emphasizing the significance of heavy-tailed parameterization in learning under complex and sparse scenarios. We also remark that HT-SPG is extremely easy to implement and train and can be integrated with any learning task endowed with complex and sparse rewards distribution for enhanced performance.

\begin{figure}[t]
	\centering
	\subfigure[Hopper-v2 of MuJoCo.]{\label{fig:hopper}		\includegraphics[width=0.32\textwidth]{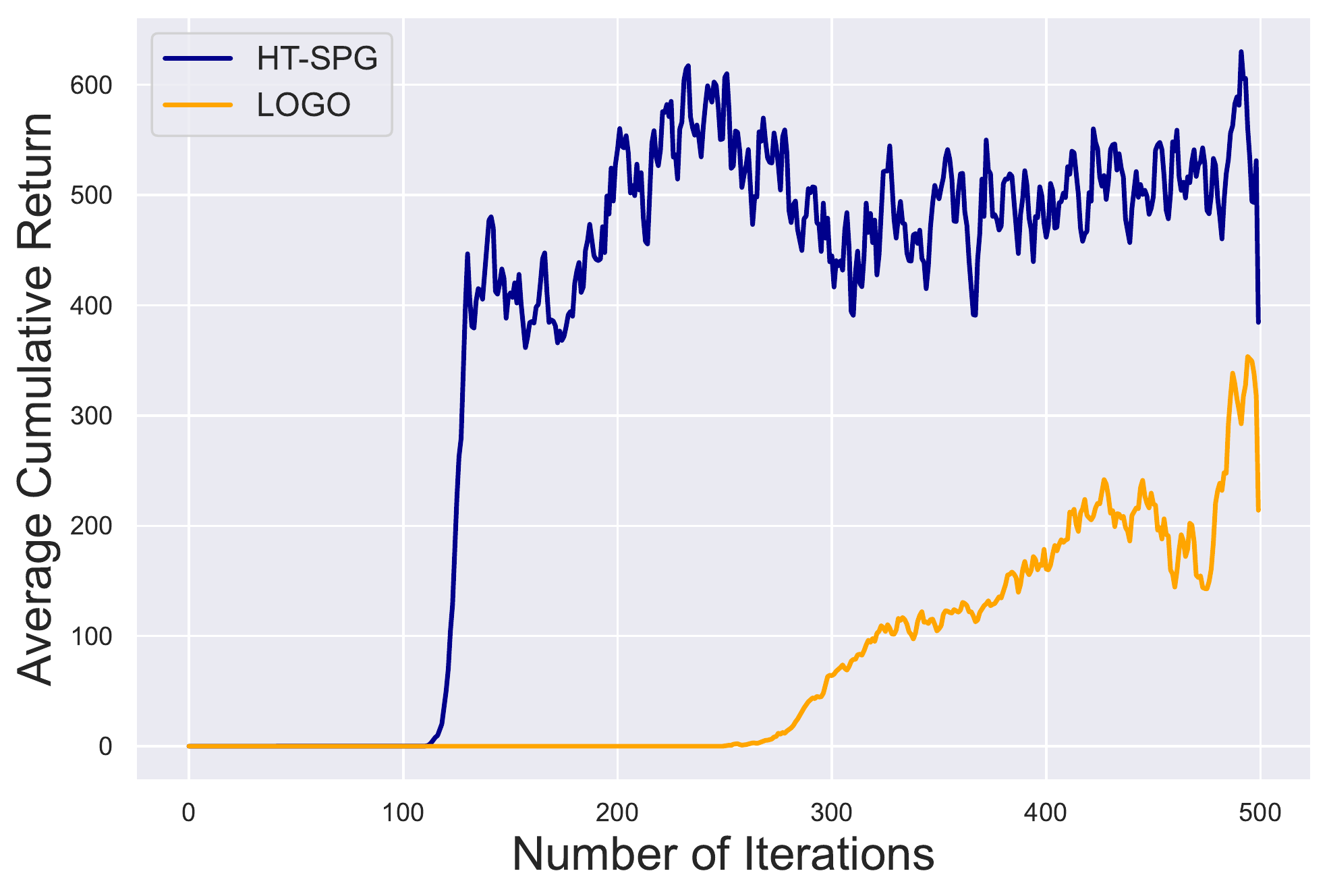}}
	%\subfigure[Pathological Mountain Car.]{\label{fig:b}		\includegraphics[width=0.4\textwidth]{figs/mario_reward_without_var.pdf}}
		
	\caption{ \normalsize We plot the average cumulative reward return of the proposed algorithm HT-SPG with the state-of-the-art LOGO algorithm. We note that heavy-tailed policy parametrization is able to induce implicit exploration into the algorithm and HT-SPG starts receiving higher rewards in almost half iterations as compared to LOGO. HT-SPG also converges to a high reward policy very quickly. 
	Note : Total number of training samples = Batch size * No of Iterations.}
		\label{mario_wt}
	\end{figure}
\subsection{MuJuCo Sparse Environments}
\begin{figure}[t]
	\centering
	\subfigure[Hopper-v2 of MuJoCo.]{\label{fig:hopper22}		\includegraphics[width=0.42\textwidth]{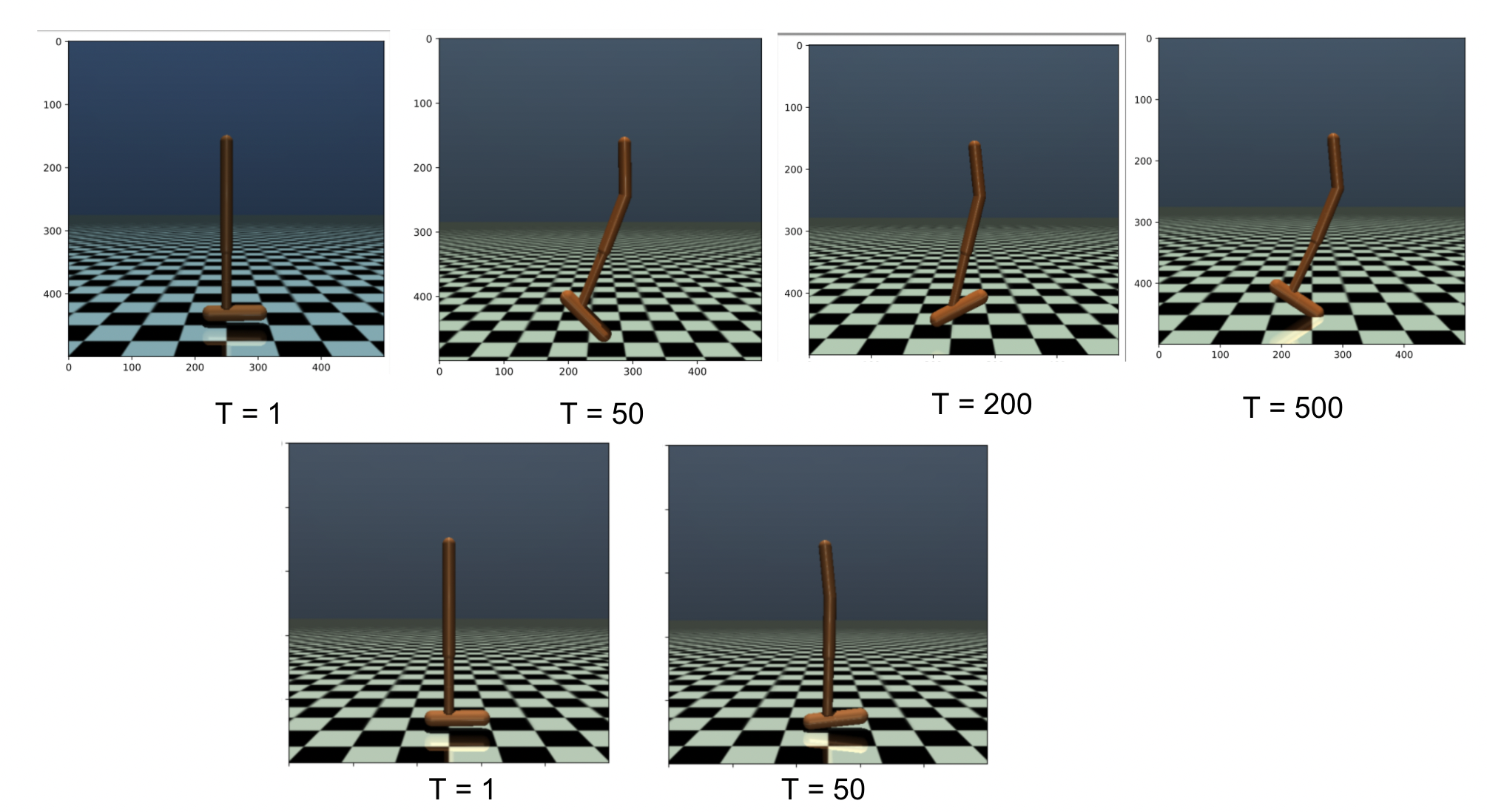}}
	%\subfigure[Pathological Mountain Car.]{\label{fig:b}		\includegraphics[width=0.4\textwidth]{figs/mario_reward_without_var.pdf}}
		
	\caption{ \normalsize We note that the proposed heavy-tailed policy (top row) was able to better learn and hop the robot for $500$ number of steps while the Gaussian policy (bottom row) could only do it for $50$ steps. Note : Model trained on the 400 iteration has been used for evaluation.}
		\label{new}
	\end{figure}

In this section, we consider the complex sparse  MuJoCo environments of Hopper (see Fig. \ref{Hopper}) and test the performance of the proposed HT-SPG algorithm. We compare it with the state-of-the-art LOGO algorithm \cite{rengarajan2022reinforcement}. The state and actions spaces for these environments are no longer scalar anymore and require us to deal with multi-variate distributions for the policy parametrizations. State-space is $12$-dimensional, action space is $3$-dimensional  linear reward for forward progress and a quadratic penalty on a joint effort to produce the reward with a bonus of $+1$ for being in a non-terminal state. The episodes when the hopper fell over, which was defined by thresholds on the torso height and angle. The sparsity in reward structure is obtained by reducing the events at which reward feedback is provided. Specifically, we provide a reward of $+1$ only after the agent moves forward over $2$ units from its initial position.  We present the performance of HT-SPG as compared to the LOGO algorithm in Fig.~\ref{fig:hopper}. Since the performance of LOGO was optimized to operate with demonstrations, we considered the same learning environment with demonstrations for the proposed HT-SPG algorithm as well. We note that the proposed algorithm is able to outperform LOGO by a significant margin and exhibit better sample efficiency. In Fig. \ref{fig:hopper22}, we present the snapshots of the behavior of final policy learned by the proposed algorithm. 
%
% \begin{itemize}
% 	\item  \textbf{Sparse Hopper-v2}:  
	
% %	\item \textbf{Sparse Walker-2d}: State space is $18$-dimensional, action space is $6$-dimensional. For the walker, we added a penalty for strong impacts of the feet against the ground to encourage a smooth walk rather than a hopping gait.
% 	\end{itemize}

% \begin{figure*}[t]
% 	\centering
% 	\subfigure[???]{\label{fig:b}		\includegraphics[width=0.5\columnwidth]{figs/mountain_car_reward.pdf}}
% 	\subfigure[???]{\label{fig:b}		\includegraphics[width=0.5\columnwidth]{figs/mountain_car_without_var.pdf}}
% 	\caption{(a)  (b)}
% 		\label{fig2}
% 	\end{figure*}
	
% \begin{figure*}[t]
% 	\centering
% 	\subfigure[???]{\label{fig:b}		\includegraphics[width=0.5\columnwidth]{figs/pendulum_reward.pdf}}
% 	\subfigure[???]{\label{fig:b}		\includegraphics[width=0.5\columnwidth]{figs/pendulum_reward_without_var.pdf}}
% 	\caption{(a)  (b)}
% 		\label{fig2}
% 	\end{figure*}

%%%%%%%%%%%%%%%%%%%%%%%%%%%%%%%%%%%%%%%%%%%%%%%%%%%%%%%%%%%%%%%%%%
%%%   S   E   C   T   I   O   N   %%%%%%%%%%%%%%%%%%%%%%%%%%%%%%%%
%%%%%%%%%%%%%%%%%%%%%%%%%%%%%%%%%%%%%%%%%%%%%%%%%%%%%%%%%%%%%%%%%%
%%%%%%%%%%%%%%%%%%%%%%%%%%%%%%%%%%%%%%%%%%%%%%%%%%%%%%%%%%%%%%%%%%%%%%%%%%%%%%%%
\section{Conclusion, Limitations, and Future Work
} \label{sec:conclusion}
In this work, we proposed a novel approach to deal with sparse reward in continuous control robotics task. Instead of relying on reward shaping or seeking information from expert's demonstrations, we utilize heavy-tailed policy parametrizations along with momentum based gradient tracking to learn in sparse robotics environments. We prove the efficacy of the proposed ideas on various robotics tasks including inverted pendulum of OpenAI Gym and Hopper-v2 of MuCoCo environments. The main limitation of the current approach is that we cannot prove any theoretical convergence guarantees for the proposed approach. For future work, it would be interesting to look at the sample complexity analysis of the proposed HT-PSG. 

%%%%%%%%% REFERENCES
{\small
	\bibliographystyle{IEEEtran}
\bibliography{RL_2}
}

\end{document}